%% file: main.tex
\documentclass[ ]{ceurart} 
\usepackage[utf8]{inputenc}
\usepackage{enumitem}
\usepackage[most]{tcolorbox}
\usepackage[colorinlistoftodos,prependcaption,textsize=tiny]{todonotes}

\input{sections/commands}

\begin{document}

\copyrightyear{2022}
\copyrightclause{Copyright for this paper by its authors.
  Use permitted under Creative Commons License Attribution 4.0
  International (CC BY 4.0).}

\conference{ALTNLP2022: The International Conference on Agglutinative Language Technologies as a challenge of Natural Language Processing, June 6-8, 2022, Koper, Slovenia}

\title{BoAT v2 - A Web-Based Dependency Annotation Tool with Focus on Agglutinative Languages}

\author[1]{Salih Furkan Akkurt}[%
email=furkan.akkurt@boun.edu.tr
]

\author[2]{Büşra Marşan}[%
email=busra.marsan@boun.edu.tr
]

\author[1]{Susan Uskudarli}[%
email=suzan.uskudarli@boun.edu.tr
]

\address[1]{ Department of Computer Engineering, Boğaziçi University, İstanbul, Turkey }
\address[2]{ Department of Linguistics, Boğaziçi University, İstanbul, Turkey }

\input{sections/abstract}

\begin{keywords}
natural language processing \sep
linguistic annotation \sep
annotation tool \sep
web application \sep
dependency parsing \sep
Universal Dependencies
\end{keywords}

\maketitle

\input{sections/introduction}
\input{sections/boatvone}
\input{sections/related-work}
\input{sections/requirements-design}
\input{sections/implementation}
\input{sections/use-case}
\input{sections/discussion-future-work}

\begin{acknowledgments}
This work was supported by Boğaziçi University Research Fund Grant Number 16909.
\end{acknowledgments}

\bibliography{main}

\newpage
\appendix
\input{sections/appendix}

\end{document}

%% file: sections/commands.tex
\newcommand{\conllu}{\textsc{c}o\textsc{nll-u}}

\newcommand{\boatvone}{\textsc{b}o\textsc{at}-{\scriptsize v1}}
\newcommand{\boatvtwo}{\textsc{b}o\textsc{at}-{\scriptsize v2}}
\newcommand{\bountreebank}{\textsc{boun} \textsc{T}reebank}

\newcommand{\id}{\textsc{id}}
\newcommand{\form}{\textsc{form}}
\newcommand{\lemma}{\textsc{lemma}}
\newcommand{\upos}{\textsc{upos}}
\newcommand{\xpos}{\textsc{xpos}}
\newcommand{\feats}{\textsc{feats}}
\newcommand{\head}{\textsc{head}}
\newcommand{\deprel}{\textsc{deprel}}
\newcommand{\deps}{\textsc{deps}}
\newcommand{\misc}{\textsc{misc}}

\newcommand{\ud}{\textsc{ud}}

%% file: sections/abstract.tex
\begin{abstract}
    The value of quality treebanks is steadily increasing due to the crucial role they play in the development of natural language processing tools.
    The creation of such treebanks is enormously labor-intensive and time-consuming.
    Especially when the size of treebanks is considered, tools that support the annotation process are essential.
    Various annotation tools have been proposed, however, they are often not suitable for agglutinative languages such as Turkish.
    \boatvone\ was developed for annotating dependency relations and was subsequently used to create the manually annotated \bountreebank\ (UD\_Turkish-BOUN).
    In this work, we report on the design and implementation of a dependency annotation tool (\boatvtwo) based on the experiences gained from the use of \boatvone, which revealed several opportunities for improvement.
    \boatvtwo\ is a multi-user and web-based dependency annotation tool that is designed with a focus on the annotator user experience to yield valid annotations.
    The main objectives of the tool are to: (1) support creating valid and consistent annotations with increased speed, (2) significantly improve the user experience of the annotator, (3) support collaboration among annotators, and (4) provide an open-source and easily deployable web-based annotation tool with a flexible application programming interface (API) to benefit the scientific community.
    This paper discusses the requirements elicitation, design, and implementation of \boatvtwo\ along with examples.
\end{abstract}

%% file: sections/introduction.tex
\section{Introduction}
\label{sec:introduction}

Treebanks are important resources in the development of natural language processing (NLP) tools.
Quality NLP tools need treebanks that are manually annotated by linguistic experts.
This is especially true for agglutinative languages due to their complex morphologies.
The creation of such treebanks is highly labor-intensive and time-consuming due to the meticulous attention required.
Thus, annotation tools that support this process are essential.

In recent years, there have been significant efforts to bridge the gap in data resources available for agglutinative low-resource languages.
Dependency annotation involves annotating each token of a sentence with linguistically relevant values.
Universal Dependencies (UD)~\cite{UD} is the most widely accepted standard for dependency annotations.
The dependency annotation format of \ud\ called \conllu\ (Computational Natural Language Learning) defines a set of linguistic tags for annotation purposes.
For example, Universal part-of-speech (\upos) tag is used for annotating the part-of-speech of a token.
The tag for morphological features (\feats) is used for annotating additional lexical and grammatical properties of tokens which are not covered by other tags.
Specifically for agglutinative languages, the \feats\ tag is very frequently annotated with multiple values due to the complex morphology of such languages.
Thus, the effort required to annotate agglutinative languages is significantly higher.
Annotation tools with drag-drop and mouse-based interfaces, although appealing, are not well suited for agglutinative languages as they require alternating among input modalities, disrupting the flow.

\boatvone~\cite{turk2021resources} is an annotation tool that was developed to support dependency annotation of morphologically rich languages (MRLs) to produce treebanks compliant with the \ud\ framework~\cite{UD}.
The experience during the use of it revealed several points of improvement for such annotation tools.
The main takeaway was a much better understanding of the time, effort, cognitive load, and extra information requirements of the annotation process.
Improvements regarding these aspects should, consequently, produce higher quality data resources.

This work presents a web-based collaborative dependency annotation tool (\boatvtwo) that has been designed based on the experience with \boatvone.
In light of the feedback from the experience, we wanted the tool to be a web application that supports multiple users to enable a collaborative environment for annotations.
Several user experience improvements were implemented to enhance the flow of an annotation session.
The design and implementation of the tool aimed to: (1) support creating valid and consistent annotations with increased speed, (2) significantly improve the user experience of the annotator, (3) allow collaboration among annotators during the annotation process, and (4) provide an open-source and easily deployable web-based annotation tool with an API to benefit the scientific community.
The development started with requirements elicitation, for which earlier experiences and in-depth interviews with an experienced annotator were taken into account.

The current prototype is being evaluated with positive feedback.
This feedback indicates that the ability to collaborate within the tool is needed to increase the efficiency of multi-annotator treebank creation.
The final version will be made available on Boğaziçi University's NLP platform~\cite{DIP} and provided as an open-source resource.

The main contributions of this work are:
\begin{itemize}
\setlength\itemsep{0em}
        \item Design of a dependency annotation tool based on requirements elicited from experienced annotators who are linguists,
        \item The development of a tool that takes into account the annotator experience to improve resulting annotations,
        \item Multi-user support to provide individual spaces for annotations, computation of inter-annotator agreements, and other potential collaboration,
        \item The development of a web-based annotation tool based on a supporting API for programmatic access and extensibility, and
        \item Packaging of the tool to support easy access by virtualizing it using Docker~\cite{docker} and providing the code as an open-source resource.
\end{itemize}

The remainder of this paper is organized as follows:
Section~\ref{sec:boatvone} presents \boatvone,
Section~\ref{sec:related} presents related work,
Section~\ref{sec:requirements} describes the requirements and design,
Section~\ref{sec:implementation} presents the implementation and new features,
Section~\ref{sec:annotation} presents a use case of annotation, and
Section~\ref{sec:discussion} provides a discussion along with our conclusions.

%% file: sections/boatvone.tex
\section{BoAT-v1}
\label{sec:boatvone}

\boatvone~\cite{turk-etal-2019-turkish} is a standalone tool for annotating treebanks that is compatible with the \ud\ framework~\cite{UD} that is implemented using Python~\cite{python} and Qt~\cite{qt}.
It was specifically developed for annotating Turkish treebanks and is particularly suitable for agglutinative languages, however it can be used for other languages.
It supports the annotation of a single treebank at a time.
The annotations are stored in a file in \conllu\ format.
The file is updated during the annotation process.
It uses a validation script developed by \ud\ to display errors.
\boatvone\ was used to create the \bountreebank~\cite{turk2021resources,UD-Boun-Treebank} -- a manually annotated Turkish dependency treebank comprising close to 10 thousand sentences.

% main features of v1
For each sentence to be annotated, the annotator is shown a table which has a token per row with its corresponding tags (\id, \form, \lemma, \upos, \xpos, \feats, \head, \deprel, \deps, and \misc\ as detailed in~\cite{turk-etal-2019-turkish}).
The annotator manually enters values for each tag of each token.
It supports the splitting and joining of lemmas which is particularly significant for agglutinative languages.
Tokens that are split result in additional rows for each part.
Figure~\ref{fig:anno-fig-v1} in Appendix~\ref{sec:appendix-a} shows a sentence that is being annotated.
The token ``yoktu'' (ID: 4-5) is split into ``yok'' (ID: 4) and ``tu'' (ID: 5). 
Furthermore, it parses the \feats\ tag's value into individual morphological features.
The number of morphological features are much higher in agglutinative languages, therefore the value of the \feats\ tag often consists of numerous values.
As such, they are difficult to read.
\boatvone\ supports viewing these features individually under their associated features (Case=Nom|Number=Sing|Person=3 can be shown in columns ``Case'', ``Number'', and ``Person'' with the values ``Nom'', ``Sing'', and ``3'').
It also allows the annotators to be able to take notes for specific annotations.

%% file: sections/related-work.tex
\section{Related Work}
\label{sec:related}

Annotation tools may be characterized in terms of their accessibility and the support they provide for languages, various annotation categories, user interface modalities, standards, and multiple annotators.
Adherence to standards is recommended to get the most benefit from the annotated data sources.
Universal Dependencies~\cite{UD} is an actively growing standard that intends to cover all languages and its support for agglutinative languages is evolving.

Several dependency annotation tools have been proposed such as \textit{brat}~\cite{brat}, \textit{UD Annotatrix}~\cite{tyers-etal:2018} and \textit{DgAnnotator}~\cite{dgannotator,UD-tools}.
These tools are not developed for a specific language, but can be used with a variety of languages.
They mostly rely on mouse-based user interaction, which is inefficient for annotating agglutinative languages due to the need for extensive annotation for most tokens.

The ITU Treebank Annotation Tool~\cite{pamay-etal-2015-annotation} was developed for Turkish.
It was written in Java as an open-source standalone tool and has several versions.
It has three stages of annotation: morphological analysis, morphological disambiguation and syntax analysis.
It offers semi-automated support for annotators through analyzers for creating new datasets as well as correcting already existing Turkish treebanks.
It mostly relies on mouse-based interactions and doesn't support the \ud\ framework.
This tool has been used to annotate the ITU Web Treebank~\cite{itu-web-tb}.

WebAnno~\cite{webanno} is a web-based open-source annotation tool, that is not restricted to dependency annotations but has support for morphological, syntactical, and semantic annotations also, with multi-user support.
To annotate features of a token, it requires several mouse clicks which is impractical for MRLs.
The sentences to be annotated are displayed consecutively, unlike many tools that focus on a single sentence at a time.

\boatvtwo\ is the second iteration of \boatvone.
It improves on it by including a search functionality, a database to represent sentences in a more essential way rather than a plaintext file and an accessible web interface with an API for flexibility.
\boatvtwo\ has been developed to reduce clutter that \boatvone\ was found to have in some spaces by the feedback of \boatvone.

%% file: sections/requirements-design.tex
\section{Requirements and Design}
\label{sec:requirements}

Requirements elicitation and validation meetings were held with annotators who worked on the \bountreebank\ involving thousands of sentences to identify the software requirements.
The main requirements that emerged are:
\begin{itemize}[before=\normalfont, font=\itshape, align=left,noitemsep,topsep=0pt,parsep=3pt,partopsep=0pt,labelsep=3pt,align=left]
    \item[Collaborative annotation:]
        The sheer size of treebanks necessitates that they are annotated via multiple annotators.
        The tool must support multiple annotators working on the same treebank.
        The ability to share annotation experiences is vital for reference and consistency.
        Hence, the tool must provide means for registering multiple annotators and keeping track of their respective annotations.
    \item[Search:]
        While annotating a sentence, annotators may need to refer to previous annotations for guidance.
        An annotator should be able to search a treebank according to surface forms, and importantly, according to linguistic features.
        They should be able to perform complex searches that involve several criteria such as \ud\ tags, individual features of the \feats\ tag, and surface forms.
        This requirement reduces the cognitive load as well as supports consistency among annotations.
    \item[Keyboard-oriented input:]
        Annotations of agglutinative languages require annotation of several features for individual tokens in comparison to analytical languages like English.
        While drag-and-drop interfaces can work well for analytical languages, they don't for MRLs as they require more toggles between mouse and keyboard interactions which is frustrating.
        Keyboard-based interaction must be offered for all possible tasks.
    \item[Support for Sentence annotation:]
        To support the speed and accurate annotations, \textit{autocompletion} should be supported whenever possible.
        Due to typologies of and syncretism being generally higher in MRLs, automatic parsing of such treebanks is difficult and they more often than not fail to create valid parsings of tokens.
        Thus, refining automatic parsing is important for agglutinative languages, which requires splitting of lemmas.
        Words in MRLs tend to have morphemes stacked on roots.
        Annotators must be able to refine/correct automatically parsed entries, which for agglutinative languages includes \textit{splitting of lemmas}.
    \item[Use of screen real estate and customization:]
        Annotators annotate numerous sentences every session.
        For agglutinative languages, sentences tend to be long and complicated.
        The view of such sentences can cover a considerable part of the screen.
        The annotator must focus on the sentence being annotated, the lemmas, and all the features, which requires much concentration.
        The annotation process must not involve scrolling, if possible.
        For very long sentences, this may not be possible; however, the sentence being annotated should never go out of vision.
        Furthermore, each annotator has their unique preferences for how they annotate a sentence.
        They must have some control regarding showing or hiding certain information according to their preferences.
        Overall efficient use of the screen real estate is important to convey the context needed to annotate long sentences.
\end{itemize}

Given these requirements, we decided that a web-based application that supports multiple authenticated users would support a collaborative platform for annotators.
The treebank and user annotations will persist in a database which makes managing the data and searches much more reliable and efficient.
Also, an API is developed to support flexibility and extensibility.
Finally, in addition to making the tool open-source, we containerized the application using Docker\cite{docker} to support its accessibility.

%% file: sections/implementation.tex
\section{Implementation}
\label{sec:implementation}

The annotation tool is implemented using Python~\cite{python}, the web application development framework \textit{Django}~\cite{django} and the API framework \textit{Django REST Framework} (DRF)~\cite{drf}. The webpages use Bootstrap~\cite{bootstrap}.
PostgreSQL~\cite{psql} is used as a database management system.
The models reflect the \ud\ format of sentences.
Most of the sentence annotation functionalities are similar to \boatvone.
User input is validated according to \ud\ and errors are reported on the annotation page.
Three alternative forms of dependency graph visualizations are supported, two of which are newly added and compact horizontal to reduce the required screen real estate~\cite{spacy,spyssalo}.

The following features have been implemented to support the creation of valid annotations with increased speed in a collaborative manner:
\begin{itemize}[before=\normalfont, font=\itshape, align=left,noitemsep,topsep=0pt,parsep=3pt,partopsep=0pt,labelsep=3pt,align=left]
    \item[Treebank handling:]
        The tool should support the annotation of multiple treebanks.
        \boatvtwo\ uses a database to persist the annotations of multiple annotators of multiple treebanks.

    \item[Sentence annotation:]
    	An annotator selects the sentence they want to annotate.
        The sentence annotation page is very similar to \boatvone.
        It consists of three main parts: (1) A table with rows for every token in the sentence and columns, which represent the \ud\ tags of a sentence, corresponding to their annotations; (2) the dependency graph of the sentence; and (3) errors from validation according to the \ud\ framework.
        The dependency graph and errors are synchronized with the annotations.
        Several dependency graph presentations are supported to suit the annotator's preference.
        Vertical graphs can consume a significant amount of screen real estate, which can lead to loss of focus in long sentences common in agglutinative languages.

        An annotator may need to stop the annotation of a sentence for some reason (i.e. complexity or external interruption).
        To capture the state of an annotation, a status is introduced in \boatvtwo\ whose values may be: ``New", ``Draft" and ``Complete".
        The status of a sentence that has not been annotated is ``New''.
	    An annotator can set the status to ``Draft'' or ``Complete''.

        The annotator is able to perform almost all operations, more than what \boatvone\ allowed, via keyboard action, based on the demand of the annotators (see Section~\ref{sec:requirements} for details).
        Upon the experience of annotating a Turkish treebank with \boatvone, the annotators have reported being very pleased with the convenience and speed resulting from keyboard-based interaction.

    \item[Improved searching for reference and consistency:]
        A search functionality is introduced in \boatvtwo.
        Users are able to search for previously annotated sentences in combinations of surface text, \ud\ tags, and features.
		Without a good search feature, an annotator would have to manually search the \conllu\ file for relevant cases (e.g. how to annotate some surface form's \upos\ tag).
		The surface form might have been inconsistently annotated, alas it is unlikely that a manual search would reveal this case.
		In such a case, the annotator would likely use the first encountered as a reference.
        The situation gets more complicated for syncretic morphemes such as \textit{-ki} in Turkish.
        For example, in the sentence ``Evdeki halılar yıkandı.'' (\textit{The rugs at home were washed.}), the \textit{-ki} acts as an adjectivizer.
        However, in ``Benim halılarım yün, Ayşeninkiler sentetik.'' (\textit{My rugs are woolen. Ayşe's are synthetic.}), it is pronominal.
        Searching for sentences where the \textit{-ki} morpheme occurs via text search would be hopeless as there would be too many hits since they occur very frequently.

        To facilitate effective searching, we have implemented search functionality based on combinations of text and \ud\ tags.
        Regular expression-based search is also supported.
        This feature supports annotators to share experiences, which consequently is expected to result in more accurate and consistent treebanks.

    \item[Inter-annotator agreement:]
        The consistency of annotations among annotators is an indicator of the quality of the resulting resource.
        Inter-annotator agreement computes the consistency among annotators.
        Since this tool keeps track of annotator actions, unlike its predecessor, such computations are straightforward.
        Some visualizations shall accompany these statistics.

\end{itemize}

%% file: sections/use-case.tex
\section{Using BoAT-v2}
\label{sec:annotation}

\begin{figure}[th!]
    \centering
    \tcbox[left=0mm,right=0mm,top=0mm,bottom=0mm,boxsep=1pt,arc=0mm,boxrule=0.5pt,colframe=black,sharp corners]
    {\includegraphics[width=1\textwidth]{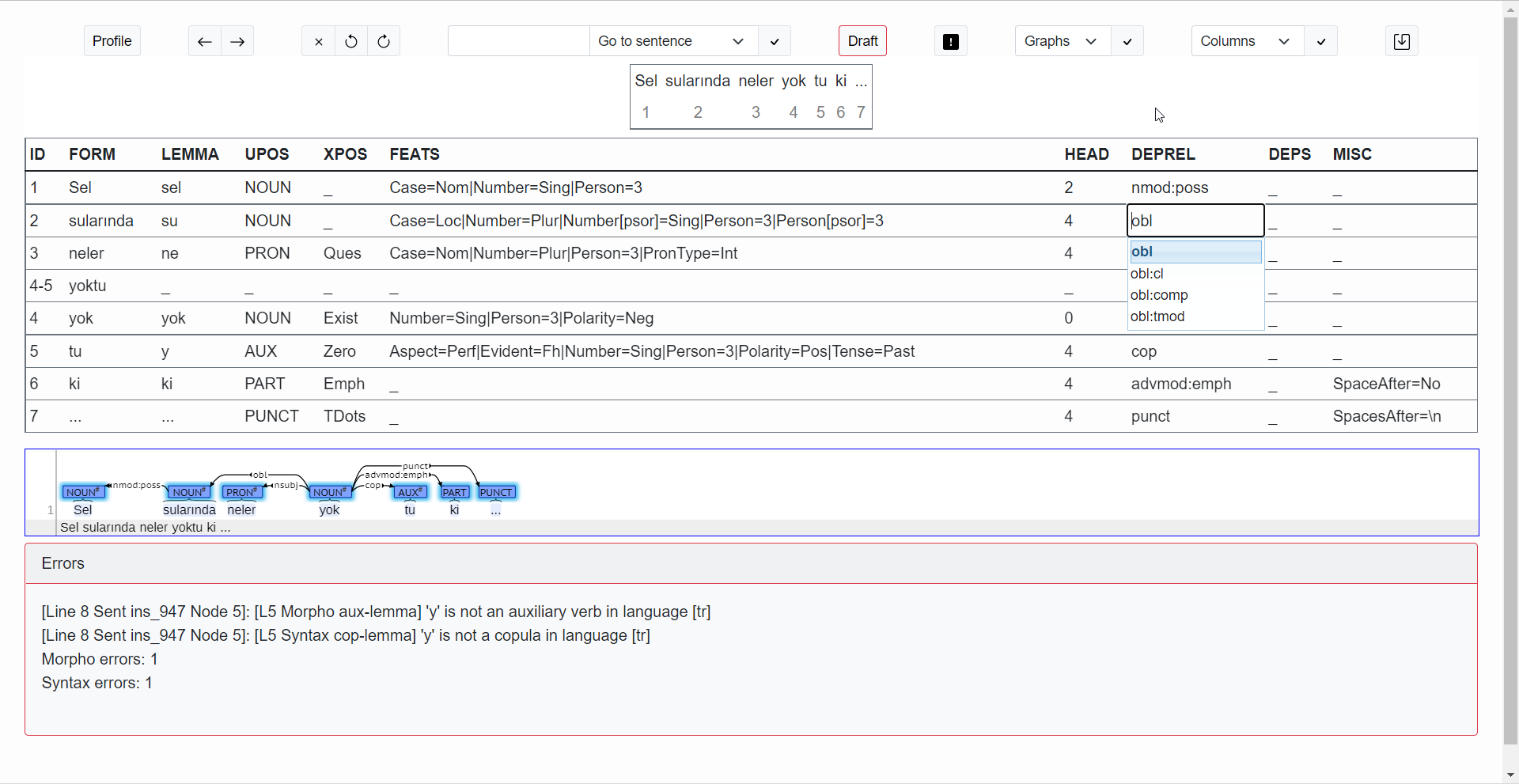}}
    \caption{The annotation screen captured while the sentence ``Sel sularında neler yoktu ki...'' is being annotated. The \deprel\ tag for the surface form ``sularında'' is being annotated by selecting among the valid alternatives that appear in the pop-up. Selections can be made with the use of arrow keys.}
    \label{fig:anno-fig}
\end{figure}

A typical annotation can proceed as follows:
An annotator selects a sentence from a treebank.
An annotation table appears with the sentence parsed according to the \ud\ format.
Each row corresponds to a token and its annotations.
Figure~\ref{fig:anno-fig} shows the annotation view while an annotator is annotating a Turkish sentence ``Sel sularında neler yoktu ki...'' (translation: \textit{What wasn't in the flood waters...}).

An annotator can make use of dependency graphs, errors, and search during annotation.
Dependency graphs are visual cues for how lemmas are dependent upon one another.
Errors are helpful reminders compliant with \ud.

Annotators can customize the columns and dependency graph in accordance with their preferences.
When an annotation is finished, its status can be set to ``Complete''.
The annotator can search for previously made annotations in combinations of text and feature values to refer to previously made annotations.

%% file: sections/discussion-future-work.tex
\section{Discussion and Conclusions}
\label{sec:discussion}

\boatvtwo\ aims to extend the functionality of \boatvone\ as a collaborative web-based application to support the annotation process based on previous experiences.
We developed a web application that supports agglutinative languages as described in Section~\ref{sec:implementation}.

The implementation choices served our goals well.
We believe that having experts in linguistics and experienced annotators in agglutinative treebank creation was instrumental in understanding the requirements and the design process.
We held numerous elicitation interviews and further meetings for clarifications and feedback requests.

We used modern software development tools and management practices during the development lifecycle of this tool.
The development of an API enables various extensions of this tool and access to the treebanks.
The containerization with Docker~\cite{docker} has facilitated easy delivery and deployment.
It will be made available on Boğaziçi University's NLP platform~\cite{DIP} as a demo as well as an open-source resource.
The availability of the tool as a user as well as a developer is valuable for future use and developments.

This tool is in the testing phase and has had encouraging early feedback.
We compared the annotation of several sentences with the same number of tokens using \boatvone\ and \boatvtwo.
While we kept the number of words in a sentence constant, we did not use the same sentences since having previously annotated a sentence would impact the annotation on another version.
Keeping the number of words the same provides a somewhat comparable experience.
There was a noticeable speedup (approximately 30\%) using \boatvtwo.
Among the new features that are most appreciated are autocompletion, condensed dependency tree representation, significant reduction on scrolling, keyword search, and search by morphological features.
The non-search related features are instrumental to retaining focus.

We are encouraged by the early responses to this tool and anticipate its extensions.
In fact, our implementation of \boatvtwo\ resulted in a revision request for \boatvone\ to include the focus enhancing features.
This also resulted in significant speedup and improved user experience, which was reported as a qualitative observation by an annotator.
Presently, our testing is focused on more extensive cases (annotation of sentences with varying degrees of complexity) and, importantly, the multi-user functionalities.
For this purpose, we are in the process of recruiting several annotators with a background in linguistics.

%% file: sections/appendix.tex
\section{BoAT-v1}
\label{sec:appendix-a}

\begin{figure}[th!]
    \centering
    \tcbox[left=0mm,right=0mm,top=0mm,bottom=0mm,boxsep=1pt,arc=0mm,boxrule=0.5pt,colframe=black,sharp corners]
    {\includegraphics[width=1\textwidth]{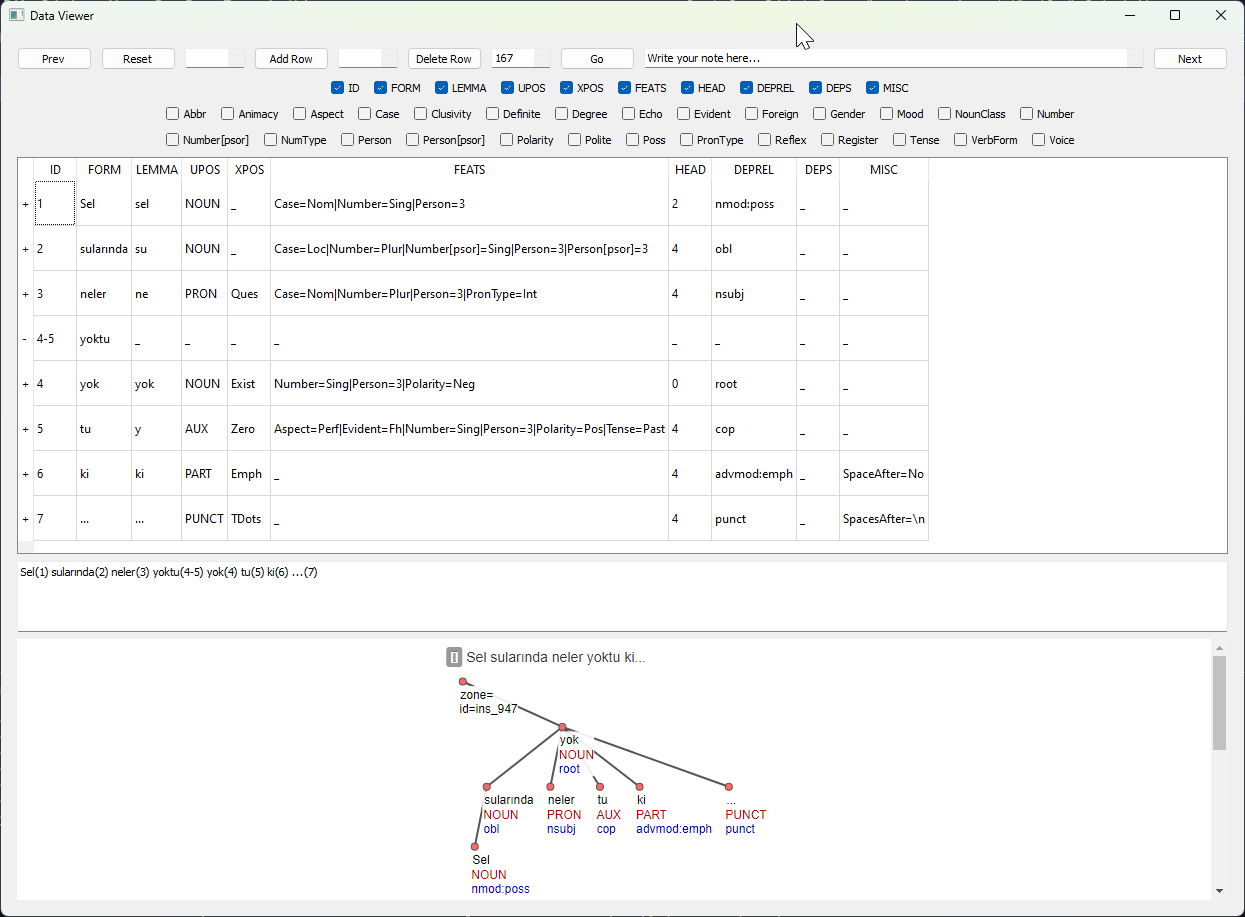}}
    \caption{The annotation screen on \boatvone, captured while the sentence ``Sel sularında neler yoktu ki...'' is being annotated.}
    \label{fig:anno-fig-v1}
\end{figure}

%% file: main.bbl
\begin{thebibliography}{21}
\expandafter\ifx\csname natexlab\endcsname\relax\def\natexlab#1{#1}\fi
\providecommand{\url}[1]{\texttt{#1}}
\providecommand{\href}[2]{#2}
\providecommand{\path}[1]{#1}
\providecommand{\DOIprefix}{doi:}
\providecommand{\ArXivprefix}{arXiv:}
\providecommand{\URLprefix}{URL: }
\providecommand{\Pubmedprefix}{pmid:}
\providecommand{\doi}[1]{\href{http://dx.doi.org/#1}{\path{#1}}}
\providecommand{\Pubmed}[1]{\href{pmid:#1}{\path{#1}}}
\providecommand{\bibinfo}[2]{#2}
\ifx\xfnm\relax \def\xfnm[#1]{\unskip,\space#1}\fi
%Type = Misc
\bibitem[{{Universal Dependencies}(2022)}]{UD}
\bibinfo{author}{{Universal Dependencies}}, \bibinfo{title}{{UD}},
  \bibinfo{year}{2022}. \URLprefix \url{https://universaldependencies.org},
  \bibinfo{note}{[Online; last accessed 2 May 2022]}.
%Type = Article
\bibitem[{T{\"u}rk et~al.(2021)T{\"u}rk, Atmaca, {\"O}zate{\c{s}}, Berk, Bedir,
  K{\"o}ksal, Ba{\c{s}}aran, G{\"u}ng{\"o}r, and
  {\"O}zg{\"u}r}]{turk2021resources}
\bibinfo{author}{U.~T{\"u}rk}, \bibinfo{author}{F.~Atmaca},
  \bibinfo{author}{{\c{S}}.~B. {\"O}zate{\c{s}}}, \bibinfo{author}{G.~Berk},
  \bibinfo{author}{S.~T. Bedir}, \bibinfo{author}{A.~K{\"o}ksal},
  \bibinfo{author}{B.~{\"O}. Ba{\c{s}}aran},
  \bibinfo{author}{T.~G{\"u}ng{\"o}r}, \bibinfo{author}{A.~{\"O}zg{\"u}r},
\newblock \bibinfo{title}{Resources for turkish dependency parsing: Introducing
  the boun treebank and the boat annotation tool},
\newblock \bibinfo{journal}{Language Resources and Evaluation}
  (\bibinfo{year}{2021}) \bibinfo{pages}{1--49}.
%Type = Misc
\bibitem[{{Boğaziçi University}(2022)}]{DIP}
\bibinfo{author}{{Boğaziçi University}}, \bibinfo{title}{{TABILAB Repository
  Home}}, \bibinfo{year}{2022}. \URLprefix \url{https://nlp.cmpe.boun.edu.tr},
  \bibinfo{note}{[Online; last accessed 24 April 2022]}.
%Type = Misc
\bibitem[{{Docker Inc.}(2022)}]{docker}
\bibinfo{author}{{Docker Inc.}}, \bibinfo{title}{{Home - Docker}},
  \bibinfo{year}{2022}. \URLprefix \url{https://www.docker.com},
  \bibinfo{note}{[Online; last accessed 2 May 2022]}.
%Type = Inproceedings
\bibitem[{T{\"u}rk et~al.(2019)T{\"u}rk, Atmaca, {\"O}zate{\c{s}}, K{\"o}ksal,
  Ozturk~Basaran, Gungor, and {\"O}zg{\"u}r}]{turk-etal-2019-turkish}
\bibinfo{author}{U.~T{\"u}rk}, \bibinfo{author}{F.~Atmaca},
  \bibinfo{author}{{\c{S}}.~B. {\"O}zate{\c{s}}},
  \bibinfo{author}{A.~K{\"o}ksal}, \bibinfo{author}{B.~Ozturk~Basaran},
  \bibinfo{author}{T.~Gungor}, \bibinfo{author}{A.~{\"O}zg{\"u}r},
\newblock \bibinfo{title}{{T}urkish treebanking: Unifying and constructing
  efforts},
\newblock in: \bibinfo{booktitle}{Proceedings of the 13th Linguistic Annotation
  Workshop}, \bibinfo{publisher}{Association for Computational Linguistics},
  \bibinfo{address}{Florence, Italy}, \bibinfo{year}{2019}, pp.
  \bibinfo{pages}{166--177}. \URLprefix
  \url{https://www.aclweb.org/anthology/W19-4019}.
  \DOIprefix\doi{10.18653/v1/W19-4019}.
%Type = Misc
\bibitem[{{Python Software Foundation}(2022)}]{python}
\bibinfo{author}{{Python Software Foundation}}, \bibinfo{title}{Welcome to
  python.org}, \bibinfo{year}{2022}. \URLprefix \url{https://www.python.org},
  \bibinfo{note}{[Online; last accessed 24 April 2022]}.
%Type = Misc
\bibitem[{{The Qt Company}(2022)}]{qt}
\bibinfo{author}{{The Qt Company}}, \bibinfo{title}{{Qt | Cross-platform
  software development for embedded \& desktop}}, \bibinfo{year}{2022}.
  \URLprefix \url{https://www.qt.io/}, \bibinfo{note}{[Online; last accessed 25
  May 2022]}.
%Type = Misc
\bibitem[{T{\"{u}}rk et~al.(2020)T{\"{u}}rk, Atmaca, and {\c{S}aziye
  Bet{\"{u}}l {\"{O}}zate{\c{s}} and G{\"{o}}zde Berk and Seyyit Talha Bedir
  and Abdullatif K{\"{o}}ksal and Balk{\i}z {\"{O}}zt{\"{u}}rk Ba{\c{s}}aran
  and Tunga G{\"{u}}ng{\"{o}}r and Arzucan
  {\"{O}}zg{\"{u}}r}}]{UD-Boun-Treebank}
\bibinfo{author}{U.~T{\"{u}}rk}, \bibinfo{author}{F.~Atmaca},
  \bibinfo{author}{{\c{S}aziye Bet{\"{u}}l {\"{O}}zate{\c{s}} and G{\"{o}}zde
  Berk and Seyyit Talha Bedir and Abdullatif K{\"{o}}ksal and Balk{\i}z
  {\"{O}}zt{\"{u}}rk Ba{\c{s}}aran and Tunga G{\"{u}}ng{\"{o}}r and Arzucan
  {\"{O}}zg{\"{u}}r}}, \bibinfo{title}{{UD Turkish BOUN}},
  \bibinfo{year}{2020}. \URLprefix \url{https://universaldependencies.org/tr},
  \bibinfo{note}{[Online; last accessed 27 March 2022]}.
%Type = Inproceedings
\bibitem[{Stenetorp et~al.(2012)Stenetorp, Pyysalo, Topi\'{c}, Ohta, Ananiadou,
  and Tsujii}]{brat}
\bibinfo{author}{P.~Stenetorp}, \bibinfo{author}{S.~Pyysalo},
  \bibinfo{author}{G.~Topi\'{c}}, \bibinfo{author}{T.~Ohta},
  \bibinfo{author}{S.~Ananiadou}, \bibinfo{author}{J.~Tsujii},
\newblock \bibinfo{title}{{brat}: a web-based tool for {NLP}-assisted text
  annotation},
\newblock in: \bibinfo{booktitle}{Proceedings of the Demonstrations Session at
  {EACL} 2012}, \bibinfo{publisher}{Association for Computational Linguistics},
  \bibinfo{address}{Avignon, France}, \bibinfo{year}{2012}.
%Type = Inproceedings
\bibitem[{Tyers et~al.(2018)Tyers, Sheyanova, and Washington}]{tyers-etal:2018}
\bibinfo{author}{F.~M. Tyers}, \bibinfo{author}{M.~Sheyanova},
  \bibinfo{author}{J.~N. Washington},
\newblock \bibinfo{title}{{UD Annotatrix: An annotation tool for Universal
  Dependencies}},
\newblock in: \bibinfo{booktitle}{Proceedings of the 16th International
  Workshop on Treebanks and Linguistic Theories (TLT16)}, \bibinfo{year}{2018},
  pp. \bibinfo{pages}{10--17}.
%Type = Misc
\bibitem[{Attardi(2022)}]{dgannotator}
\bibinfo{author}{G.~Attardi}, \bibinfo{title}{{DgAnnotator}},
  \bibinfo{year}{2022}. \URLprefix
  \url{http://medialab.di.unipi.it/Project/QA/Parser/DgAnnotator/},
  \bibinfo{note}{[Online; last accessed 27 March 2022]}.
%Type = Misc
\bibitem[{{Universal Dependencies}(2022)}]{UD-tools}
\bibinfo{author}{{Universal Dependencies}}, \bibinfo{title}{{UD tools}},
  \bibinfo{year}{2022}. \URLprefix
  \url{https://universaldependencies.org/tools}, \bibinfo{note}{[Online; last
  accessed 22 April 2022]}.
%Type = Inproceedings
\bibitem[{Pamay et~al.(2015)Pamay, Sulubacak, Toruno{\u{g}}lu-Selamet, and
  Eryi{\u{g}}it}]{pamay-etal-2015-annotation}
\bibinfo{author}{T.~Pamay}, \bibinfo{author}{U.~Sulubacak},
  \bibinfo{author}{D.~Toruno{\u{g}}lu-Selamet},
  \bibinfo{author}{G.~Eryi{\u{g}}it},
\newblock \bibinfo{title}{The annotation process of the {ITU} web treebank},
\newblock in: \bibinfo{booktitle}{Proceedings of The 9th Linguistic Annotation
  Workshop}, \bibinfo{publisher}{Association for Computational Linguistics},
  \bibinfo{address}{Denver, Colorado, USA}, \bibinfo{year}{2015}, pp.
  \bibinfo{pages}{95--101}. \URLprefix \url{https://aclanthology.org/W15-1610}.
  \DOIprefix\doi{10.3115/v1/W15-1610}.
%Type = Inproceedings
\bibitem[{Torunoğlu-Selamet et~al.(2015)Torunoğlu-Selamet, Pamay, Sulubacak,
  and Eryiğit}]{itu-web-tb}
\bibinfo{author}{D.~Torunoğlu-Selamet}, \bibinfo{author}{T.~Pamay},
  \bibinfo{author}{U.~Sulubacak}, \bibinfo{author}{G.~Eryiğit},
\newblock \bibinfo{title}{The annotation process of the itu web treebank},
\newblock \bibinfo{year}{2015}. \DOIprefix\doi{10.3115/v1/W15-1610}.
%Type = Misc
\bibitem[{{WebAnno}(2022)}]{webanno}
\bibinfo{author}{{WebAnno}}, \bibinfo{title}{{WebAnno - Documentation}},
  \bibinfo{year}{2022}. \URLprefix
  \url{https://webanno.github.io/webanno/documentation},
  \bibinfo{note}{[Online; last accessed 5 May 2022]}.
%Type = Misc
\bibitem[{{Django Software Foundation}(2022)}]{django}
\bibinfo{author}{{Django Software Foundation}}, \bibinfo{title}{Django
  documentation}, \bibinfo{year}{2022}. \URLprefix
  \url{https://docs.djangoproject.com/en/4.0}, \bibinfo{note}{[Online; last
  accessed 25 March 2022]}.
%Type = Misc
\bibitem[{{Encode OSS Ltd.}(2022)}]{drf}
\bibinfo{author}{{Encode OSS Ltd.}}, \bibinfo{title}{{Django REST framework}},
  \bibinfo{year}{2022}. \URLprefix \url{https://www.django-rest-framework.org},
  \bibinfo{note}{[Online; last accessed 27 March 2022]}.
%Type = Misc
\bibitem[{{Bootstrap}(2022)}]{bootstrap}
\bibinfo{author}{{Bootstrap}}, \bibinfo{title}{{Bootstrap v5.1 Documentation}},
  \bibinfo{year}{2022}. \URLprefix \url{https://getbootstrap.com/docs/5.1},
  \bibinfo{note}{[Online; last accessed 5 May 2022]}.
%Type = Misc
\bibitem[{{The PostgreSQL Global Development Group}(2022)}]{psql}
\bibinfo{author}{{The PostgreSQL Global Development Group}},
  \bibinfo{title}{{PostgreSQL}}, \bibinfo{year}{2022}. \URLprefix
  \url{https://www.postgresql.org/docs}, \bibinfo{note}{[Online; last accessed
  27 March 2022]}.
%Type = Misc
\bibitem[{Explosion(2022)}]{spacy}
\bibinfo{author}{Explosion}, \bibinfo{title}{{spaCy}}, \bibinfo{year}{2022}.
  \URLprefix \url{https://spacy.io}, \bibinfo{note}{[Online; last accessed 27
  March 2022]}.
%Type = Misc
\bibitem[{Pyysalo(2022)}]{spyssalo}
\bibinfo{author}{S.~Pyysalo}, \bibinfo{title}{conllu.js}, \bibinfo{year}{2022}.
  \URLprefix \url{http://spyysalo.github.io/conllu.js}, \bibinfo{note}{[Online;
  last accessed 25 March 2022]}.

\end{thebibliography}
